\def\year{2020}\relax
\documentclass[letterpaper]{article} 
\usepackage{aaai20}  
\usepackage{times}  
\usepackage{helvet} 
\usepackage{courier}  
\usepackage[hyphens]{url}  
\usepackage{graphicx} 
\usepackage{graphicx}  
\usepackage[toc,page]{appendix}
\usepackage{subcaption}
\usepackage{caption}
\usepackage[dvipsnames]{xcolor}

\usepackage[newfloat,frozencache,cachedir=.]{minted}

\usepackage{bm,array}
\usepackage{mathtools}
\usepackage{xcolor}

\title{The Animal-AI Environment:\\Training and Testing Animal-Like Artificial Cognition}
\author{
Benjamin Beyret,\textsuperscript{\rm 1,4} 
Jos{\'e} Hern{\'a}ndez-Orallo,\textsuperscript{\rm 2,4} 
Lucy Cheke,\textsuperscript{\rm 3,4} 
Marta Halina,\textsuperscript{\rm 3,4} \\
{\bf \Large 
Murray Shanahan,\textsuperscript{\rm 1,4} 
Matthew Crosby\textsuperscript{\rm 1,4}
} \\  
\textsuperscript{\rm 1}Imperial College London, UK \hspace{10pt}
\textsuperscript{\rm 2}Universitat Polit{\`e}cnica de Val{\`e}ncia, Spain \\
\textsuperscript{\rm 3}University of Cambridge, UK  \hspace{10pt}
\textsuperscript{\rm 4}Leverhulme Centre for the Future of Intelligence, UK\\
bb1010@ic.ac.uk, jorallo@upv.es, \{lgc23, mh801\}@cam.ac.uk, 
\{m.shanahan, m.crosby\}@imperial.ac.uk
}

\newenvironment{yaml}{\captionsetup{type=listing}}{}
\SetupFloatingEnvironment{listing}{name=Config.}
\nocopyright

\begin{document}
\maketitle

\begin{abstract}
Recent advances in artificial intelligence have been strongly driven by the use of game environments for training and evaluating agents. Games are often accessible and versatile, with well-defined state-transitions and goals allowing for intensive training and experimentation. However, agents trained in a particular environment are usually tested on the same or slightly varied distributions, and solutions do not necessarily imply any understanding. If we want AI systems that can model and understand their environment, we need environments that explicitly test for this. Inspired by the extensive literature on animal cognition, we present an environment that keeps all the positive elements of standard gaming environments, but is explicitly designed for the testing of animal-like artificial cognition. All source-code is publicly available (see appendix).\end{abstract}

\section{Introduction}

From chess \cite{silver_2017} to Go \cite{silver_2016}, from older Atari games \cite{bellemare_2012} to Starcraft 2 \cite{vinyals_2017}, we have seen a wide variety of challenging environments where AI now outperforms humans. Successes such as these in deep reinforcement learning (DRL) have been driven by the introduction of game environments and physics simulators \cite{todorov_2012}, and have even resulted in the  transfer of trained agents to the real world for robotic manipulations \cite{openai_2018}. These impressive results are only a first step towards agents that can robustly interact with their environments. In these settings, most tasks used for training are identical, or extremely similar, to the ones used for testing. Whilst the agent has not always been exposed to the test tasks during training, it has probably been exposed to tasks drawn from the same distribution. This leads to agents that reach superhuman performance on specific problems, but are unable to generalise \cite{packer2018assessing}.

AI systems can solve and outperform humans in many complex tasks, but, whilst progress is ongoing, a {\em single} agent cannot compare to a human's ability to adapt and generalise \cite{espeholt2018impala}. Humans are no longer the best Go players, but they are still the best at functioning across a wide number of environments and adapting to unexpected situations. Even non-human animals exhibit the ability to solve a wider array of tasks than AI systems are currently capable of \cite{herrmann2007humans}.

In animal cognition research, ethologists observe organisms in their natural environments whilst comparative psychologists design carefully-controlled studies in order to probe animals' cognitive and behavioural capacities \cite{thorndike1911animal}. Over the last century the testing process has been refined to eliminate confounding factors, noise and other non-cognitive elements that are interfere with identifying the targeted skill \cite{shaw2017cognitive}. A wide range of standardised tests have been developed, along with some well-known experimental paradigms (radial mazes, landmark arenas, etc.). With these tools, comparative psychologists can adapt and customise tasks to fit a wide range of species, finding the best way to measure a skill or clarify some particular question about behaviour.

Following the above-mentioned limitations of current AI testbeds and the theoretical and methodological tools available in comparative psychology, we introduce the Animal-AI environment, which is used in the Animal-AI Olympics competition, to test trained agents on tasks inspired by animal cognition research. The environment includes all the beneficial parts of gaming environments: it has a simple reward function, action space and complex yet deterministic state transition function based on a simulated physics engine. It is also set up to test  multiple cognitive abilities in a single agent. The environment consists of a small arena with seven different classes of objects that can be placed inside. These include reward objects (corresponding to food) and building blocks which can be combined to create complex cognitive tests.

The paper is structured as follows. We first present current benchmarks and related environments and identify a need for cognitive behavioural testing. We then introduce experimental paradigms from comparative cognition and show how they can be adapted for AI. Following this, we present the Animal-AI environment, illustrating how it can be used as a training and testing platform. Finally, we introduce the Animal-AI testbed, which is used to test a wide range of cognitive skills in the Animal-AI Olympics competition.

\section{Related Benchmarks in AI}


Progress in DRL in recent years has been fuelled by the use of games and game-inspired simulated environments  \cite{castelvecchi2016tech,hernandez2017new}. In these settings, for an agent or policy to be successful it must integrate perception of the scene (usually in 2D or 3D) with the right choice of action sequences that allow the agent to achieve the goal conditions. Some important benchmarks are simply collections of existing games, such as the very popular \textit{Arcade Learning Environment} (ALE) \cite{bellemare2015arcade,machado2018revisiting}, with dozens of (2D) Atari 2600 games. In a similar vein, OpenAI Gym \cite{brockman2016openai} provides a common interface to a collection of RL tasks including both 2D and 3D games. As machine learning methods conquer more games, there seem to be endless new (and old) games ready to be put forward as the next challenge  \cite{jaderberg2018human}. 

Moving towards generalisation, we see games designed to include variations or set stages, where some skill transfer between training and testing is necessary. {\em CoinRun} \cite{cobbe2018quantifying} is a 2D arcade-style game aimed at generalisation and transfer. It is designed to be simple enough so that individual levels are easy to solve when trained on, but the task is to create an agent that can solve unseen variations. Another example, {\em Obstacle Tower} \cite{juliani2019obstacle} is a new 3D game based on Montezuma's revenge, one of the harder (for AI) Atari games, whose stages are generated in a procedural way. The procedural generation ensures that the agent can be tested on unseen instances. 

While the above collections and game-like environments give the possibility of adding new tasks or task variations, other platforms are designed to be customisable to allow for a wide variety of challenges. For instance, using the video game definition language (VGDL), new real-time 2D games can be created relatively easily. This has led to several \textit{General Video Game AI} (GVGAI) competitions, with new games for each edition \cite{perez20162014}. 
{\em ViZDoom} \cite{kempka2016vizdoom} is a research platform with customisable scenarios based on the 1993 first-person shooting video game Doom that has been used to make advancements in model-based DRL \cite{ha2018world}.
Microsfoft's {\em Malmo} \cite{johnson2016malmo}, which is based on the block-based world of Minecraft, also makes it possible to create new tasks, ranging from navigation and survival to collaboration and problem solving. 
 Finally, {\em DeepMind Lab} \cite{beattie2016deepmind} is an extensible 3D platform with simulated real-world physics built upon id Software’s Quake III Arena. DeepMind Lab is primarily set up for large navigational problems or tests based on visual cues. In contrast, the Animal-AI environment is useful for setting up small navigation problems, and also supports interactive experiments that test an agent's ability to reason about the physical environment based on its (simulated) physics, like those used in comparative psychology.

Without being comprehensive (as new platforms appear very regularly), this AI-evaluation landscape shows some common characteristics. (1) Most traditional platforms are devised to set new challenges assuming some incremental improvement over current AI technology. (2) To be considered a real challenge, oversimplified tasks that look like toy problems are avoided (even if AI cannot currently solve them).  
(3) When many tasks are integrated into a benchmark, it is not always the same trained agent (but the same algorithm) that is evaluated on them, with retraining for each task. (4) Only a few more recent platforms are aimed towards generalisation (of the same agent) where the training and the test sets differ --but not always \mbox{substantially--} and retraining is not allowed. (5) Most challenges are task-oriented instead of ability-oriented \cite{hernandez2016aire}, without explicitly identifying the skill to be measured. 
We challenge (1)-(5), introducing an environment that presents a new paradigm for AI. We use tests that are simple in the context of comparative psychology, but are challenging for AI and include many ability-oriented tasks aimed at testing generalisation of a single agent.

Another ability-oriented approach is \emph{bsuite}, which presents a series of reinforcement learning tasks designed to be easily scalable and to provide a measure for a number of core capabilities 
\cite{osband2019}. These tests are deliberately simple to allow for a more accurate measure of the ability being tested. In contrast, our approach, inspired by biological intelligence, focuses on measuring cognitive skills scaffolded by an agent's ability to perceive, navigate, and interact with its environment. Our tests are also deliberately simple to a human observer, but because they are built up from perception, they are much more complicated to solve.


\section{Animal Cognition Benchmarks}

\begin{figure}[t] 
    \centering
    \begin{subfigure}[b]{0.22\textwidth}
        \includegraphics[width=\textwidth]{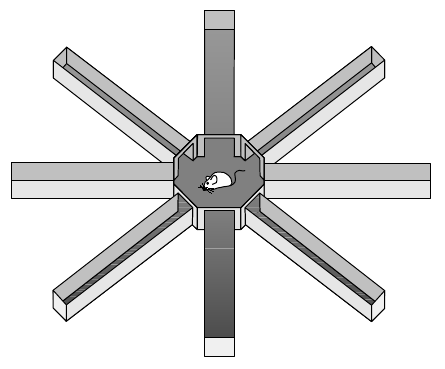}
        \caption{An eight-arm radial maze, used for spatial memory in rats.}
        \label{fig:radial-a}
    \end{subfigure}
    ~
    \begin{subfigure}[b]{0.22\textwidth}
        \includegraphics[width=\textwidth]{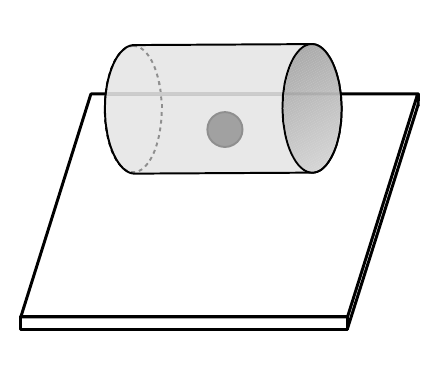}
        \caption{Detour task \cite{kabadayi2018detour}.
}
        \label{fig:cylinder}
    \end{subfigure}
    \caption{Two different apparatus commonly used in comparative cognition. Food is placed in the arms of the maze for (\ref{fig:radial-a}) and inside a tube for (\ref{fig:cylinder}). Whether or not the animal collects food, and more importanlty how it does it, demonstrates various skills in animals.}
    \label{fig:animalcogexperiments}
\end{figure}

The study of animal behaviour includes areas such as ethology, behavioural ecology, evolutionary psychology, comparative psychology and more recently, comparative cognition \cite{wasserman2006comparative,shettleworth2009cognition}. 
In the latter two, animals are usually evaluated in carefully designed and controlled conditions, using standardised procedures. 
Radial arm mazes, for instance, are frequently employed to study spatial memory in rats (Figure \ref{fig:radial-a}). An experiment is usually constrained by three design principles: (1) Solving the task requires some basic functions that are taken for granted or have been previously demonstrated in the animal, such as being able to move, recognise shapes and colours, desire rewards, etc. (2) Beyond these basic functions, the cognitive skill or capacity under investigation (e.g., spatial memory) is required to perform successfully on the task. (3) All other confounding factors are eliminated or controlled for, such as cues that would allow subjects to solve the task (e.g., odors in a maze) without relying on the target skill. Other experimental confounds, such as personality traits or motivation, are also avoided or taken into account in the statistical analysis of the results \cite{shaw2017cognitive}. Note that these three principles make experimental tasks used in comparative cognition very different from the AI testbeds reviewed in the previous section. 

An experiment is then designed to analyse or evaluate a particular cognitive function. For instance, the task might be to find food in an eight-arm maze (fig. \ref{fig:radial-a}), and the way in which participants succeed or fail in the task provides insight into their cognitive abilities. If an animal retraces its steps down previously explored arms, for example, then this suggests it is not exhibiting spatial memory compared to an animal that can solve the maze systematically. The three principles above are combined with a strict protocol for evaluation. Trials can also vary in difficulty (e.g., number of food items in the maze, set of objects and their arrangement, time for each trial or between them) and presentation (e.g., training episodes). Recently, so-called ``mini test batteries" \cite{shaw2017cognitive} are becoming popular in comparative cognition. They include a small number of different tasks that target a particular function (e.g., spatial memory), standardised across species.



\section{AI Evaluation Inspired by Animal Cognition}

The route for measuring cognitive functions (or functional domains) through mini test batteries and combining them to analyse more complex patterns or profiles of behaviour (such as general intelligence) is a powerful methodology that can be extended (with the appropriate adaptations) for the evaluation of AI systems. 
%
Consequently, we introduce the Animal-AI environment as a new paradigm for training and testing AI algorithms. We provide a platform which allows researchers to design experimentation protocols in a way similar to animal cognition research. 
As discussed above, when testing animals for cognitive skills it is customary 
that the animal has to move around and interact with  a set of objects in order to retrieve one or more pieces of food. 

In order to make experiments comparable between the Animal-AI environment and the 
animal cognition literature, we mimic the way an experimenter would build a testing environment for an animal. In practice this means that the arena must follow the following principles: (1) the agent is enclosed in a fixed size arena with relatively simple objects so that we exclude as many confounding factors as possible, (2) physics must be realistic and visually reproduce how objects behave in the real world (e.g., gravity, collisions, friction, etc.). In addition, from an engineering standpoint we would expect that (3) reproducing experiments from animal cognition in our environment should be simple and fast and our environment must integrate easily with standard ML paradigms.  


We designed the environment following these principles. The agent itself is a simple sphere that can only ``walk" forward and backward and turn left and right. It cannot jump nor fly, but can climb on objects that present a slope such as ramps and cylinders. In order to simplify the problem from an AI point of view, the agent only has monocular vision through pixel inputs, and to partially mimic an animal's interoceptive abilities, it also has a sense of its own speed in its local reference frame. These simplifying assumptions are motivated in part by the principle of avoiding confounding factors originating from complex bodies or large sets of actions, as in other platforms. 

Unlike with animals, we cannot assume that AI systems are prebuilt with an understanding of intuitive physics or the ability to manipulate objects. Even if they can navigate basic versions of the environment, we cannot assume that artificial systems can avoid obstacles or push objects, which is a fundamental component of many animal cognition tests. We therefore define a playground where these affordances are common and provide this as a training arena so that agents can bring with them capabilities similar to animals.

\section{Environment Design and Specification}

The environment is provided as an executable that can be interfaced with using the provided Python API, or manually in ``play mode'' where the user can control an agent. We built on top of Unity ML-Agents \cite{mlagents} in order to provide an extra layer of abstraction as well as additional features. The environment is composed of one or more arenas, each containing a unique agent acting independently of all other agents and which can be independently configured. In figure \ref{fig:arena_all} we present one arena and the objects the user can place in it, while the next section explain how this can be done. The supplementary material contains full technical details of the arena and objects, such as their size, weight, colour, reward value, etc.

\subsection{The Arena and Objects}

\begin{figure}[h] 
    \centering
    \begin{subfigure}[b]{0.45\textwidth}
        \includegraphics[width=\textwidth]{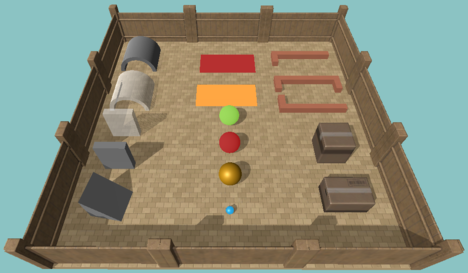}
    \end{subfigure}
\caption{Arena displaying all possible objects: agent (blue sphere at the bottom), immovable objects (left), rewards (middle) and movable objects (right).}
\label{fig:arena_all}
\end{figure}

The arena itself is made of a tiled ground and four walls represented by wooden fences, as well as an agent (blue sphere), and can hold various objects from seven categories. As shown in figure \ref{fig:arena_all}, the objects are:

\begin{itemize}
    \item {\bf Immovable objects}: these objects cannot be pushed by the agent. They separate and possibly occlude parts of the environment, allowing for the creation of different configurations that the agent has to explore to find food. They are shown on the left in figure \ref{fig:arena_all} 
    and make three categories
    (from bottom to top):
    \begin{itemize}
        \item ramps the agent can climb,
        \item walls, both opaque and transparent, that the agent must move around, and
        \item cylinders, both opaque and transparent, that the agent can move through.
    \end{itemize}
    All objects can be resized and rotated freely. They can be used individually, but can also be thought of as building blocks from which complex constructions can be built, such as those in Figure \ref{fig:animalcogexperiments}. 
    \item {\bf Reward objects}: these represent pieces of food or positive rewards. They ensure that the agent has comparable motivations to in animal cognition, where food is the goal in many experiments. We also include negative rewards, testing an agent's ability to avoid aversive stimuli. The reward objects are shown in the middle of Figure \ref{fig:arena_all} and can be divided into two categories (from bottom to top, excluding the blue agent):
    \begin{itemize}
        \item spheres that have positive (green or gold) or negative (red) rewards (green and red terminate an episode, gold does not unless it is the last positive reward item in the arena), 
        \item orange zones on the ground, \textit{hot zones}, that give negative rewards to the agent standing on it, but do not terminate an episode, and red zones,  \textit{death zones}, also give negative reward but terminate an episode and are useful for setting up no-go areas similar to those in elevated maze experiments for example \cite{pellow1985validation}. 
    \end{itemize}
    It is important that the food items are spherical. The simulated physics allows us to make use of this in the experiments.
    \item {\bf Movable objects}: Some experiments in comparative psychology require the animal to interact with objects either via grabbing, pushing or pulling, using them as {\em tools}. For simplicity, the Animal-AI agent does not have any means of holding onto objects, but it can push them by moving into them. In Figure \ref{fig:arena_all} these objects are on the right and make up the last two categories of objects (from bottom to top):
    \begin{itemize}
        \item Cardboard boxes that can be be pushed around, one heavier than the other,
        \item A set of U-shaped and L-shaped objects, all of identical weight that have different affordances to the boxes.
    \end{itemize}
\end{itemize}

\noindent On top of the physical configuration of objects in an arena, we also allow the experimenter to \textit{switch the lights on and off} as is done in some landmark navigation experiments. This can also be used to simulate experiments where items need to be intentionally obscured from view for a short period of time. With animals, this is normally done by adding a screen between the animal and the testing apparatus. In our environment the lights-off condition is implemented by replacing the visual observations sent to the agent by a black image.

\subsection{Reinforcement Learning Setup}
From a reinforcement learning perspective, for a single agent, we define observations, actions and rewards as follows (see supplementary material for more details):
\begin{itemize}
    \item \textbf{Observations}:  The agent is equipped with a single camera which shows it a pixel grid with resolution $k \times k \times 3$ where $4 \leq k \leq 512$. The agent also perceives its speed along the axes of its local referential (forward, right, up). 
    \item \textbf{Action space}: The agent can take actions $(m,t) \in \{0,1,2\}\times\{0,1,2\}$ where $m$ applies a force forward ($1$), backward ($2$) or no force ($0$), and $r$ is an instantaneous rotation to the right ($1$), left ($2$) or no rotation ($0$).
    \item \textbf{Rewards}: are obtained when a food item is collected or when the agent touches a \textit{hot} or \textit{death zone}. The agent also receives a small negative reward at each step, equal to $\frac{-1}{T}$ where $T$ is the number of timesteps in the episode (or no reward if $T=0$ which is the setting for an infinite length episode).
\end{itemize}

\noindent We offer an API to interact with the environment either as a Gym \cite{brockman2016openai} or an ML-Agents environment \cite{mlagents}.

\subsection{Task Configurations}
A crucial part of our environment is the way a user can configure tasks for both training and evaluating agents. The main way of doing so is via the use of human readable configuration files, which can precisely define the position of objects in an arena, and also allows for randomisation of most parameters (position, rotation, size, colour). The user can change the configuration of the objects in the arena between each training or testing episode via the API. This allows for agents to be trained on specifically designed task sequences, with new arenas introduced based on any user-defined trigger.

Note that with the environment, and with the Animal-AI Olympics competition that uses it, we do not place any restrictions on configurations used for training. This is a new paradigm in AI evaluation that reflects the fact that we currently do not know the most efficient setups for learning real-world skills. Part of the problem is to understand the types of environments that lead to learning of transferable skills or to find ways to learn in noisy environments without explicit direction from a reward function. For example, in our competition setup, all tests must remain secret until the end, so that no participants can train on test configurations directly.

A common configuration that has been used for training in the competition is to spawn objects randomly, increasing the number and types of objects as the agent learns. A key element for progress in Animal-AI will be to design more sophisticated training environments that allow the agent to experience a wide range of the affordances of the environment in an efficient manner and without explicit reward feedback. It is possible to build files of increasing difficulties to learn from, as in curriculum learning \cite{bengio_2009}. In the following section we show first how this might work for maze navigation alone, and then proceed to outline our complete test battery of 300 tests inspired by comparative cognition.

\section{Example: Maze Navigation}
Mazes are a simple and yet widely used problem for testing spatial memory and navigation skills of both animals \cite{olton1979mazes} and AI agents \cite{johnson2016malmo,beattie2016deepmind}. 
In the current RL paradigm, it is standard practice to train an agent in an environment that presents many configurations of a maze and then test on unseen variations drawn from the same probability distribution. This setup allows agents to solve particular mazes, but does not necessarily lead to agents that learn transferable skills that will be applicable to other types of maze.

In contrast, in animal cognition it is usually the case that we are interested in behaviour on first presentation of a particular type of maze. Training on similar mazes would be considered an experimental confound that detracts from the ability to test for navigation capabilities. To train an agent that can acquire true navigation skills we would use a set of training configurations that do not contain the types of mazes we want to test the agent on. This can easily be configured using the Animal-AI environment. 

\begin{figure}[h] 
    \centering
    \begin{subfigure}[b]{0.22\textwidth}
        \includegraphics[width=\textwidth]{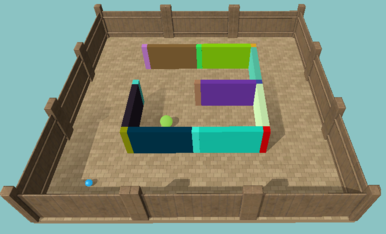}
        \caption{2x2 maze}
        \label{fig:mazes:2x2}
    \end{subfigure}
    ~
    \begin{subfigure}[b]{0.22\textwidth}
        \includegraphics[width=\textwidth]{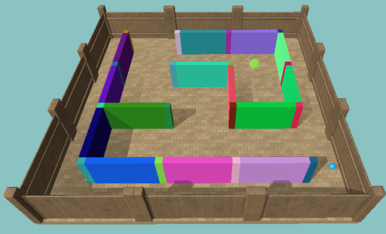}
        \caption{3x3 maze}
        \label{fig:mazes:3x3}
    \end{subfigure}
    \\
    \begin{subfigure}[b]{0.22\textwidth}
        \includegraphics[width=\textwidth]{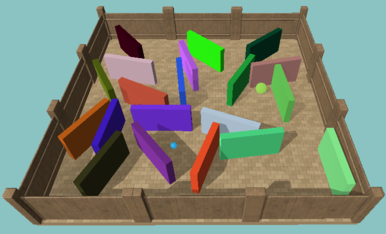}
        \caption{Scrambled maze}
        \label{fig:mazes:random}
    \end{subfigure}
    ~
    \begin{subfigure}[b]{0.22\textwidth}
        \includegraphics[width=\textwidth]{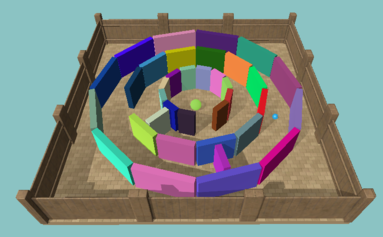}
        \caption{Circular maze}
        \label{fig:mazes:circular}
    \end{subfigure}
\caption{Samples of various types of mazes we want to evaluate an agent on.}
\label{fig:mazes}
\end{figure}

Figure \ref{fig:mazes}(a-d) shows samples of different types of mazes that we would like to evaluate a trained agent on in order to test its navigation skills. The experimenter needs to design a twofold training procedure made of a trainable agent (similar to the classic RL paradigm) and also a set of arena configurations to train the agent on. For a given algorithm, several training sets will lead to agents acquiring various skills.

\begin{figure}[h] 
    \centering
    \begin{subfigure}[b]{0.22\textwidth}
        \includegraphics[width=\textwidth]{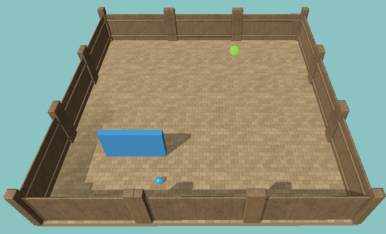}
        \caption{Curriculum level 1}
        \label{fig:curriculum:1}
    \end{subfigure}
    ~
    \begin{subfigure}[b]{0.22\textwidth}
        \includegraphics[width=\textwidth]{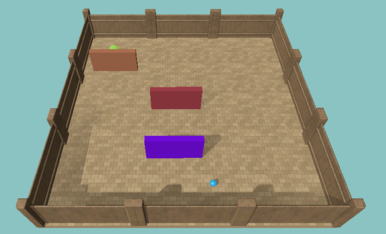}
        \caption{Curriculum level 3}
        \label{fig:curriculum:2}
    \end{subfigure}
    \\
    \begin{subfigure}[b]{0.22\textwidth}
        \includegraphics[width=\textwidth]{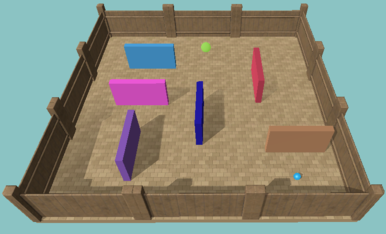}
        \caption{Curriculum level 6}
        \label{fig:curriculum:3}
    \end{subfigure}
    ~
    \begin{subfigure}[b]{0.22\textwidth}
        \includegraphics[width=\textwidth]{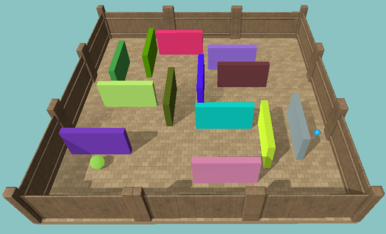}
        \caption{Curriculum level 13}
        \label{fig:curriculum:4}
    \end{subfigure}
\caption{Parts of the curriculum used to train a PPO agent to navigate around walls.}
\label{fig:curriculum}
\end{figure}

For this example, we compare two agents trained using PPO (same hyper-parameters for both, see appendix) \cite{schulman_2017}. One agent is trained to solve a randomised $2\times2$ maze (shown in Figure \ref{fig:mazes:2x2}), while the other is trained on a curriculum of arenas with increasing numbers of walls randomly placed along the $X$ and $Z$ axes, as shown in Figure \ref{fig:curriculum}. We switch the curriculum from one level of difficulty to the next when the agent reaches an $85\%$ success rate over 600 episodes. The first agent has seen maze-like configurations, with regularity and openings in the walls, but only of a single type. The second agent has not seen anything maze-like, but has experienced environments with a range of complexities.

\begin{figure}[h] 
    \centering
    \begin{subfigure}[b]{0.45\textwidth}
        \includegraphics[width=\textwidth]{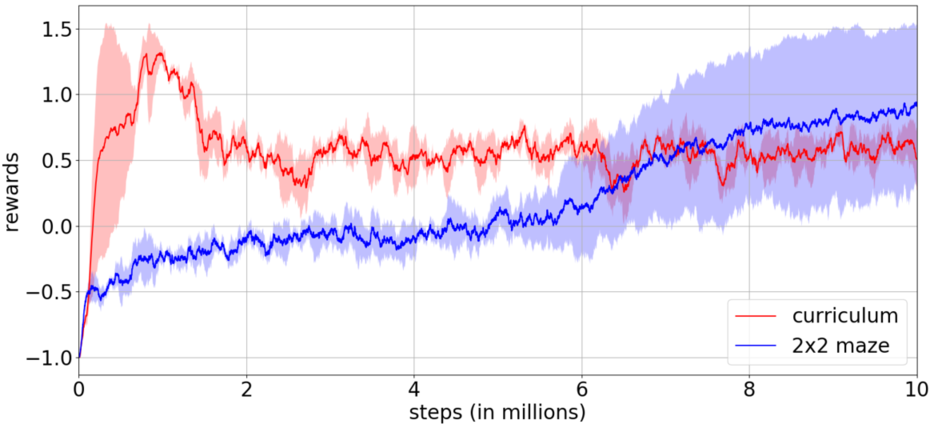}
    \end{subfigure}
\caption{Cumulative reward per episode for PPO agents trained on a curriculum (red) and a $2\times2$ maze (blue) on the training set-- mean and standard deviation over 4 random seeds. The maximum reward per episode is $2$ (obtaining no reward yields $-1$).}
\label{fig:training}
\end{figure}

\newcolumntype{C}{>{\centering\arraybackslash}p{4em}}

\begin{table*}[t]
    \begin{center}
    \begin{tabular}{|c||C|C||C|C||C|C||C|C| }
        \hline
        & \multicolumn{2}{c||}{$2\times2$ maze (fig \ref{fig:mazes:2x2})}
                & \multicolumn{2}{c||}{$3\times3$ maze (fig \ref{fig:mazes:3x3})}
                        & \multicolumn{2}{c||}{Scrambled maze (fig \ref{fig:mazes:random})}
                                & \multicolumn{2}{c|}{Circular maze (fig \ref{fig:mazes:circular})} \\
        \cline{2-9}
       &   A.R.  & S.R &   A.R.  & S.R &   A.R.  & S.R &   A.R.  & S.R  \\
        \hline
    Curriculum   &   0.70  &   64\%  &   0.92  &   70\%  &   0.4  &   53\%  &   0.03  &   39\%  \\
        \hline
    $2\times2$ maze   &   1.82  &   99\%  &   1.18  &   78\%  &   0.51  &   55\%  &   0.12  &   41\%     \\
        \hline
    \end{tabular}
    \end{center}
\caption{Average Reward (A.R.) and Success Rates (S.R.) for the two agents trained on a Curriculum and a randomised $2\times2$ maze, tested on four different types of mazes. The maximum reward an agent can obtain is equal to $2$.}
\label{results}
\end{table*}

We are interested in evaluating the two trained agents on the set of experiments defined above (Figure \ref{fig:mazes}). In Figure \ref{fig:training} we show that both agents reached equilibrium, maximising the rewards they can obtain on their two training sets respectively. Note that the red curve shows the agent learning over a curriculum, therefore the difficulty of the training tasks increase with the number of timesteps (hence the initial spike in rewards obtained). We then select the best agents: the best one overall for the agent trained on a $2\times2$ maze, and the best performer on the hardest task of the curriculum. In Table \ref{results} we show both the average rewards and the success rates of each of the two best agents, evaluated on the four types of mazes defined in Figure \ref{fig:mazes}. As expected, the agent trained on the $2\times2$ maze outperforms the other agent on the same problem and reaches 99\% accuracy. This represents a standard reinforcement learning problem and solution. 

The agent trained on $2\times2$ mazes has a significant drop in performance as we move through tests a-d, away from similarity to its training set. Even though it has seen many variations of $2\times2$ mazes, it loses considerable performance when moving to 3x3 mazes. The curriculum agent, whilst not achieving the same level of performance in the $2\times2$ maze, has much less of a drop in performance on the harder problems, achieving near parity with the standard agent on the circular maze.

These results give an idea as to how the environment can be used to train and test for animal-like cognitive skills and also how they can transfer across different problem instances. However, introducing variability into training sets adds new considerations for AI research, and also means more extensive testbeds in order to draw valid conclusions.

\section{Artificial Cognition}

Drawing conclusions about the cognitive abilities of animals requires caution, even from the most well-designed experiments. Moving to AI makes this even harder. We have seen that the Animal-AI environment introduces a new paradigm in the field of AI research by dissociating the training and testing environments, making the design of a training curriculum part of the entire training process. 
In the previous toy example we could, for example, train an agent using PPO on a training curriculum $\mathcal{A}$, and then another agent using Rainbow \cite{hessel_2017} on a training curriculum $\mathcal{B}$, yielding success rates $s_1$ and $s_2$ on the same test set. However, even if $s_1 > s_2$, we could not conclude that PPO is better than Rainbow as we cannot isolate whether the algorithm used or the training curriculum is the source of the better results. Therefore, instead of comparing training algorithms, one now needs to compare the entire package of algorithm and curriculum combined. 

The methods of comparative psychology provide us with tools for controlling confounds like the above. In order to interpret the results of a test trial in comparative cognition, one must consider a participant's evolutionary and ontogenetic history, previous training, and familiarity with the experimental setup. As noted above, when a participant performs successfully on a cognitive task, researchers carefully rule out plausible alternative explanations - ones in which the participant might have learned to solve the task using capacities other than the targeted skill. Ruling out such alternatives requires proper experimental design and statistical analysis \cite{bausman2018not}.

One partial solution to dealing with this number of confounds is to define test batteries \cite{herrmann2007humans,shaw2017cognitive} that increase the number of skills an individual agent is tested on. Compared to animals, experiments with AI systems are less costly and time intensive, so each skill can also require solving a wide variety of tests. An agent scoring well on only a few tests would likely have overfitted, or found a lucky solution. With a full test battery, an agent's successes can be compared across tests to build a profile of its behaviour and capabilities.

\section{The First Animal-AI Test Battery}

The Animal-AI test battery has been released as part of a competition alongside the Animal-AI Environment. The competition format allows us to assess performance on completely hidden tasks. If we had first released the full details of the test battery then we would have lost the opportunity to know how well AI systems could do with zero prior knowledge of the tests. Once the competition has ended we will release the full details of the tests. At this point, it will be even more important for anyone working with the test battery to disclose their training setup so that it can be determined how much they are training to solve the specific tests, and how much to learn robust food retrieval behaviour.

We designed 300 experiment configurations split over categories testing for 10 different cognitive skills. We then challenged the AI community to design both training environments and learning algorithms and submit agents that they believe display these cognitive skills. We briefly introduce the ten categories here:

\noindent \textbf{1. Basic food retrieval:} Most animals are motivated by food and this is exploited in animal cognition tests. This category tests the agent's ability to reliably retrieve food in the presence of only food items. It is necessary to make sure agents have the same motivation as animals for subsequent tests.
    
\noindent \textbf{2. Preferences:} This category tests an agent's ability to choose the most rewarding course of action. Almost all animals will display preferences for more food, or easier to obtain food, although exact details differ between species. Tests are designed to be unambiguous as to the correct course of action based on the rewards in our environment.
    
\noindent \textbf{3. Obstacles:} This category contains objects that might impede the agent's navigation. To succeed in this category the agent will have to explore its environment, a key component of animal behaviour.
    
\noindent \textbf{4. Avoidance:} This category identifies an agent's ability to detect and avoid  negative stimuli. This is a critical capacity shared by biological organisms, and is also an important prerequisite for subsequent tests.

\noindent \textbf{5. Spatial Reasoning:} This category tests an agent's ability to understand the spatial affordances of its environment. It tests knowledge of some of the simple physics by which the environment operates, and also the ability to remember previously visited locations.
    
\noindent \textbf{6. Robustness:} This category includes variations of the environment that look superficially different, but for which affordances and solutions to problems remain the same.
    
\noindent \textbf{7. Internal Models:} In these tests, the lights may turn off and the agent must remember the layout of the environment to navigate in the dark. Most tests are simple in nature, with light either alternating or only going off after some time.
    
\noindent \textbf{8. Object Permanence:} This category checks whether the agent understands that objects persist even when they are out of sight, as they do in the real world and in our environment. 
    
\noindent \textbf{9. Numerosity:} This category tests the agent's ability to make more complex decisions to ensure it gets the highest possible reward. These include situations where decisions must be made between different sets of rewards.
    
\noindent \textbf{10. Causal Reasoning:} This category tests the agent's ability  to plan ahead and contains situations where the consequences of actions need to be considered. All the tests in this category have been passed by some non-human animals, and these include some of the most striking examples of intelligence from across the animal kingdom.

By organising these problems into categories, we can get a profile for each agent, which is more informative than a simple overall test score. Creating agents that can solve such a wide range of problems is a challenging task. We expect this to require sustained effort from the AI community to solve, and that doing so will lead to many innovations in reinforcement learning.

\subsection{Current State-of-the-art}

Human performance on the tests is close to 100\%, and most of the tests have been successfully completed by some animals. Some of the easier tests are solvable by nematodes, whilst the most complex tests in the competition are only solved by animals such as corvids and chimps. Each category contains a variety of test difficulties. For example, the first tests in the basic food category spawn a single food in front of the agent whereas the final tests contain multiple moving food objects.

\begin{figure}[h]
    \centering
    \begin{subfigure}[b]{0.45\textwidth}
        \includegraphics[width=\textwidth]{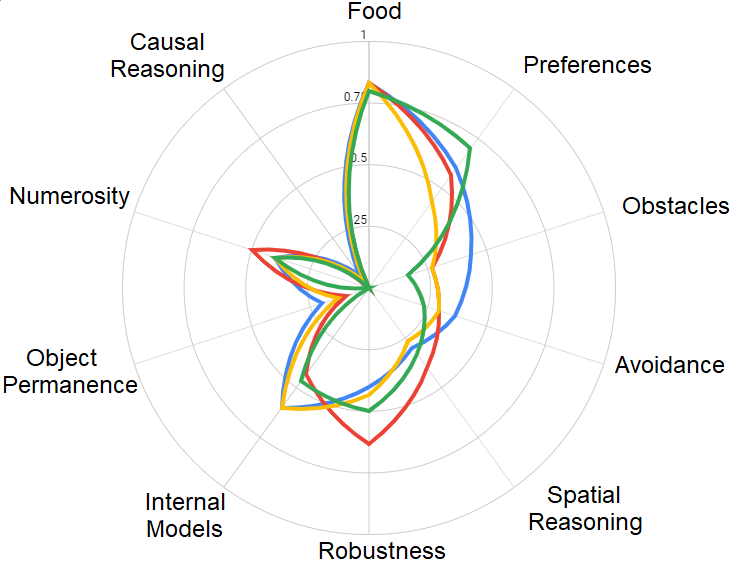}
    \end{subfigure}
\caption{Radar plot showing the performance of the top four entries to the Animal-AI Olympics at the half-way point.}
\label{fig:AAIO-results}
\end{figure}

Current state-of-the-art in AI scores around 40\%. Figure \ref{fig:AAIO-results} shows the results of the top four competitors of the competition at the moment. Methods used were all variations of reinforcement learning. We can see that certain categories, corresponding to those closer to traditional problems in reinforcement learning, are easier to solve than others. These include navigation-based problems and problems involving choosing between types of reward. On the other hand, those that require planning or that necessarily involve keeping accurate models of the world, such as for object permanence, are much harder to solve and will be an important challenge for ongoing research.

\section{Conclusions}

The Animal-AI environment is a new AI experimentation and evaluation platform that implements ideas from animal cognition in order to better train AI agents that possess cognitive skills. It is designed to contain only the necessary ingredients for performing cognitive testing built up from perception and navigation. We start with simple yet crucial tasks that many animals are able to solve. These include reward maximisation, intuitive physics, object permanence, and simple causal reasoning. By mimicking tasks and protocols from animal cognition we can better evaluate the progress of domain-general AI.

The Animal-AI platform is currently in its first iteration, and will continue to be developed as AI progresses and solves the original testbed. There are many tasks that some animals can solve that were considered too complex (for AI to solve) for the first iteration. Once it becomes possible to train agents to solve the first tasks, we will introduce more complex tests.
In future versions we also plan to include more complicated substrates and objects such as liquids, soft bodies and ropes which are common in animal experiments. A more complex physics engine would allow for embodied interactions between the agent and its environment, creating many more opportunities for animal-like tests. We will also add the possibility for multi-agent training which could allow for social interactions and social learning and also include tests that require online learning to solve correctly. 

Having said this, we expect the current testbed to remain an open challenge and provide a pathway for AI research for years to come. We have seen incredible results in game-like environments in recent years, and we hope this momentum can translate to solving the kind of problems that animals solve on a daily basis when navigating their environment or foraging for food. The Animal-AI environment is set up to make this transition as easy as possible, and to provide a series of increasingly difficult challenges towards animal-like artificial cognition. Finally, this platform helps bring the communities of AI and comparative cognition closer together, with the potential to disentangle capabilities and behaviours that are associated in animals, but can be disassociated in AI. 

\clearpage

{\fontsize{9.0pt}{10.0pt} \selectfont  
\bibliography{references.bib} 
}

\bibliographystyle{aaai}

\clearpage

\section*{Appendix}

In this appendix, we provide further details about the environment. We first describe the characteristics of the Unity-based setup, before showing how one can use the Animal-AI environment to define experiments, configure arenas and train/test agents.

\noindent All source-code is available online, along with details of the competition:

\begin{itemize}
    \item Competition details: http://animalaiolympics.com/
    \item API source-code, environment and instructions: \\ https://github.com/beyretb/AnimalAI-Olympics
    \item Environment source-code: \\ https://github.com/beyretb/AnimalAI-Environment
\end{itemize}

\subsection{Basics of the Environment}

\begin{figure}[h!] 
    \centering
    \begin{subfigure}[b]{0.22\textwidth}
        \includegraphics[width=\textwidth]{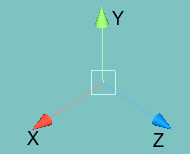}
    \end{subfigure}
    \caption{Unity referential}
    \label{fig:referential}
\end{figure}

The Animal-AI environment is built on top of Unity ML-Agents, which itself relies on the game engine Unity. We therefore use the internal physics engine to model the control and motion of the agent and the objects. The coordinate system in Unity is set as shown in figure \ref{fig:referential}, where $z$ is forward, $y$ is up and $x$ is right. A unit on each of these axes can be seen as a meter: mass is in kg and time is proportional to wall time and is machine independent (therefore velocity is uniform across machines).

\subsection{Single Arena and Agent}

\begin{figure}[h!] 
    \centering
    \begin{subfigure}[b]{0.22\textwidth}
        \includegraphics[width=\textwidth]{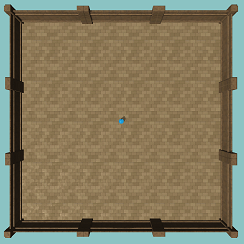}
        \caption{Empty arena (from above)}
        \label{fig:empty_arena:arena}
    \end{subfigure}
    ~
    \begin{subfigure}[b]{0.22\textwidth}
        \includegraphics[width=\textwidth]{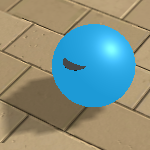}
        \caption{Agent: black dot is forward}
        \label{fig:empty_arena:agent}
    \end{subfigure}
    \caption{Basic elements of an arena}
    \label{fig:empty_arena}
\end{figure}

A single arena is made of a tiled floor and four wooden walls, whose textures never change, as depicted in figure \ref{fig:empty_arena:arena}. The arena is of size $40\times40$. Every arena also contains a single agent represented by a blue sphere of diameter $1$ and mass $1$, with a black spot to show the forward direction as shown in figure \ref{fig:empty_arena:agent}

The agent can move forward and backward by applying a force in the associated direction at each step. It can also rotate on the spot, by $\pm 6$ degrees. Once the force and the rotation have been applied, Unity's physics engine will compute the motion of the agent, managing friction, drag and collisions.

\subsection{Apparatus}

In order to allow an experimenter to design training and testing experiments in a way similar to an animal experiment, we provide a set of objects the agent can interact with.

\subsubsection{Rewards}

\begin{figure}[h!] 
    \centering
    \begin{subfigure}[b]{0.14\textwidth}
        \includegraphics[width=\textwidth]{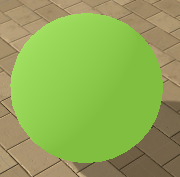}
        \caption{green food}
        \label{fig:rewards:green}
    \end{subfigure}
    ~
    \begin{subfigure}[b]{0.14\textwidth}
        \includegraphics[width=\textwidth]{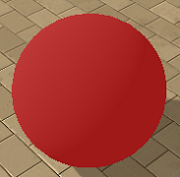}
        \caption{red food}
        \label{fig:rewards:red}
    \end{subfigure}
    ~
    \begin{subfigure}[b]{0.14\textwidth}
        \includegraphics[width=\textwidth]{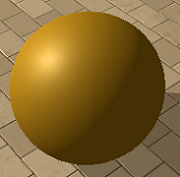}
        \caption{golden food}
        \label{fig:rewards:gold}
    \end{subfigure}
    \caption{Food objects representing rewards for the agent}
    \label{fig:rewards}
\end{figure}

As mentioned earlier, all the animal experiments we are interested in contain at least one piece of food. In our setup this is represented by three types of spheres, which bear positive or negative rewards and might or might not terminate an episode:

\begin{itemize}
    \item \textit{GoodGoal (fig \ref{fig:rewards:green})}: a positive reward represented by a green sphere of diameter $d \in [1,5]$, mass $1$, bears a reward equal to $d$. It is always green and always terminates an episode if touched by the agent.
    \item \textit{BadGoal (fig \ref{fig:rewards:red})}: a negative reward represented by a red sphere of diameter $d \in [1,5]$, mass $1$, bears a reward equal to $-d$. It is always red and always terminates an episode if touched by the agent.
    \item \textit{GoodGoalMulti (fig \ref{fig:rewards:gold})}: a positive reward represented by a golden sphere of diameter $d \in [1,5]$, mass $1$, bears a reward equal to $d$. It is always gold and will terminate an episode only if all the \textit{GoodGoalMulti} have been collected \textbf{and} there is no \textit{GoodGoal} in the arena.
\end{itemize}

All three types can also be set to move when they spawn. They are then named \textit{GoodGoalMove}, \textit{BadGoalMove} and \textit{GoodGoalMultiMove}, and will move in the direction specified by the rotation parameter.

\subsubsection{Zones}

\begin{figure}[h!] 
    \centering
    \begin{subfigure}[b]{0.22\textwidth}
        \includegraphics[width=\textwidth]{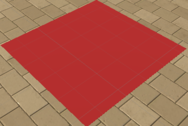}
        \caption{Death zone}
        \label{fig:zones:red}
    \end{subfigure}
    ~
    \begin{subfigure}[b]{0.22\textwidth}
        \includegraphics[width=\textwidth]{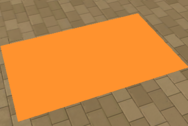}
        \caption{Hot zone}
        \label{fig:zones:orange}
    \end{subfigure}
    \caption{Zones on the ground}
    \label{fig:zones}
\end{figure}

Animal cognition experiments often include traps or holes which prevent the animal from completing the experiment, therefore ending an episode. Other experiments have used electrified areas on the ground which do not terminate an episode but will negatively reward the animal. We make use of such zones in the following way:

\begin{itemize}
    \item \textit{DeathZone (fig \ref{fig:zones:red})}: a red zone situated on the ground, of dimensions $(x,z)\in[1,40]^2$ which bears a reward of $-1$ and will terminate an episode (meant to reproduce the traps and holes of the animal experiments).
    \item \textit{HotZone (fig \ref{fig:zones:orange})}: an orange zone situated on the ground, of dimensions $(x,z)\in[1,40]^2$ which bears a reward of $min(\frac{-10}{T},-1e^{-5})$ (or $-1e^{-5}$ if $T=0$) where $T$ is the maximum number of timesteps in the current episode. It does not terminate an episode.
\end{itemize}

\subsubsection{Movable Objects}

\begin{figure}[h!] 
    \centering
    \begin{subfigure}[b]{0.22\textwidth}
        \includegraphics[width=\textwidth]{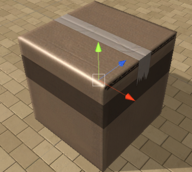}
        \caption{Light box}
        \label{fig:movable:box_light}
    \end{subfigure}
    ~
    \begin{subfigure}[b]{0.22\textwidth}
        \includegraphics[width=\textwidth]{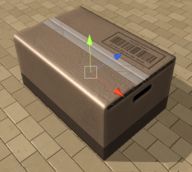}
        \caption{Heavy box}
        \label{fig:movable:box_heavy}
    \end{subfigure}
    \\
     \begin{subfigure}[b]{0.14\textwidth}
        \includegraphics[width=\textwidth]{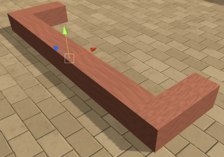}
        \caption{U object}
        \label{fig:movable:U_obj}
    \end{subfigure}
    ~
     \begin{subfigure}[b]{0.14\textwidth}
        \includegraphics[width=\textwidth]{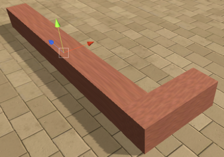}
        \caption{L object}
        \label{fig:movable:L_obj}
    \end{subfigure}
    ~
     \begin{subfigure}[b]{0.14\textwidth}
        \includegraphics[width=\textwidth]{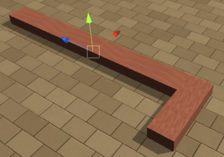}
        \caption{L object sym.}
        \label{fig:movable:L_obj2}
    \end{subfigure}
    \caption{Movable objects}
    \label{fig:movable}
\end{figure}

In order to assess the cognitive capabilities of an animal, experimenters often require this one to interact with objects by pushing, pulling or lifting various types of objects in order to obtain a reward. We provide such objects in different forms:

\begin{itemize}
    \item \textit{Cardbox1 (fig \ref{fig:movable:box_light})}: a light box make of cardboard of mass $1$ and of size $(x,y,z) \in [0.5,10]^3$
    \item \textit{Cardbox2 (fig \ref{fig:movable:box_heavy})}: a heavier box make of cardboard of mass $2$ and of size $(x,y,z) \in [0.5,10]^3$
    \item \textit{LObject (fig \ref{fig:movable:L_obj})}: a stick-like object of mass $3$ and of size $(x,y,z) \in [1,5]\times[0.3,2]\times[3,20]$
    \item \textit{LObject2 (fig \ref{fig:movable:L_obj2})}: a stick-like object of mass $3$ and of size $(x,y,z) \in [1,5]\times[0.3,2]\times[3,20]$, symetric of the \textit{LObject}
    \item \textit{UObject (fig \ref{fig:movable:U_obj})}: a stick-like object of mass $3$ and of size $(x,y,z) \in [1,5]\times[0.3,2]\times[3,20]$
\end{itemize}

\subsubsection{Immovable}

\begin{figure}[h!] 
    \centering
    \begin{subfigure}[b]{0.14\textwidth}
        \includegraphics[width=\textwidth]{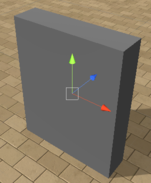}
        \caption{Opaque wall}
        \label{fig:immovable:wall}
    \end{subfigure}
    ~
    \begin{subfigure}[b]{0.14\textwidth}
        \includegraphics[width=\textwidth]{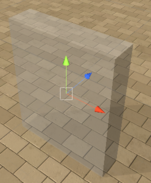}
        \caption{Transp. wall}
        \label{fig:immovable:wall_transp}
    \end{subfigure}
    ~
     \begin{subfigure}[b]{0.14\textwidth}
        \includegraphics[width=\textwidth]{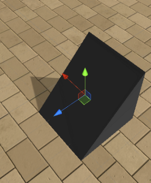}
        \caption{Ramp}
        \label{fig:immovable:ramp}
    \end{subfigure}
    \\
     \begin{subfigure}[b]{0.22\textwidth}
        \includegraphics[width=\textwidth]{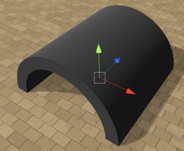}
        \caption{Opaque cylinder}
        \label{fig:immovable:cylinder}
    \end{subfigure}
    ~
     \begin{subfigure}[b]{0.22\textwidth}
        \includegraphics[width=\textwidth]{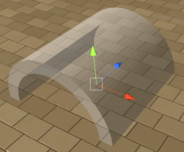}
        \caption{Transp. cylinder}
        \label{fig:immovable:cylinder_transp}
    \end{subfigure}
    \caption{Immovable objects}
    \label{fig:immovable}
\end{figure}

Finally, we are also interested in evaluating navigation and landmark detection skills which require immovable objects the animal can move around to explore its environment. We provide three types with some variations:

\begin{itemize}
    \item \textit{Wall (fig \ref{fig:immovable:wall})} a simple wall which cannot be moved by the agents and of size $(x,y,z) \in [0.1,40]\times[0.1,10]\times[0.1,40]$, its colour can be set or randomised
    \item \textit{WallTransparent (fig \ref{fig:immovable:wall_transp})} same as above but the wall is transparent and its colour cannot be set nor randomised
    \item \textit{CylinderTunnel (fig \ref{fig:immovable:cylinder})}: a tunnel the agent can move under and sometimes climb on, of size $(x,y,z) \in [2.5,10]^3$, its colour can be set or randomised
    \item \textit{CylinderTunnelTransparent (fig \ref{fig:immovable:cylinder_transp})}:same as above but the tunnel is transparent and its colour cannot be set nor randomised
    \item \textit{Ramp (fig \ref{fig:immovable:ramp}}: a ramp the agent can climb on, of size $(x,y,z) \in [0.5,40]\times[0.1,10]\times[0.5,40]$, its colour can be set or randomised
\end{itemize}

\subsection{Configuring an Arena}

\begin{figure}[h] 
    \centering
    \begin{subfigure}[b]{0.44\textwidth}
        \includegraphics[width=\textwidth]{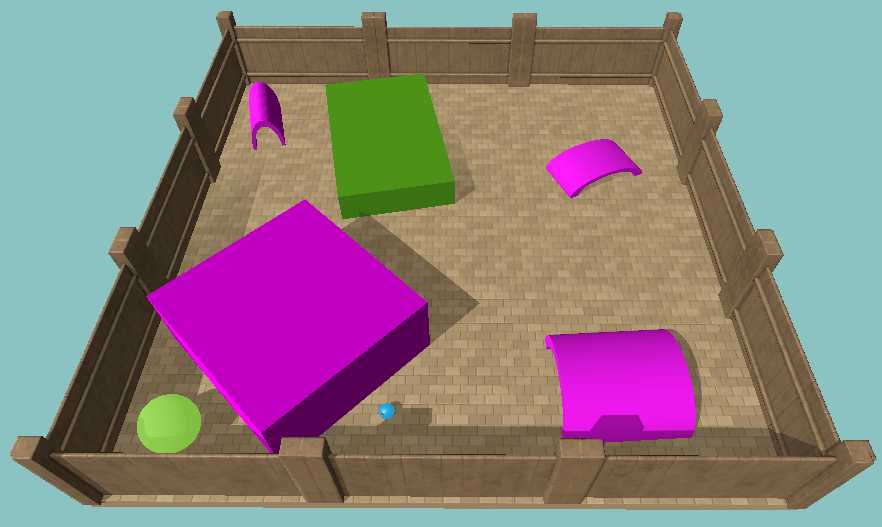}
        \caption{Configuration example}
        \label{fig:example_config}
    \end{subfigure}
\end{figure}

Each arena can be configured independently of the others. It is based on a data structure which contains a list of items to place, each item being one of those described above. We discuss in the next part exactly how the objects  are spawned (in terms of collisions and randomisation).

An arena can be configured by a single YAML file, which is human readable and easy to use. Let's walk through an example.

\begin{minipage}[]{\linewidth}
\begin{yaml}
\begin{minted}[
    gobble=4,
    frame=single
  ]{yaml}
    !ArenaConfig
    arenas:
      0: !Arena
        t: 600
        blackouts: [5,10,15,20,25]
        items:
        - !Item
          name: Wall
          positions:
          - !Vector3 {x: 10, y: 0, z: 10}
          - !Vector3 {x: -1, y: 0, z: 30}
          colors:
          - !RGB {r: 204, g: 0, b: 204 }
          rotations: [45]
          sizes:
          - !Vector3 {x: -1, y: 5, z: -1}
        - !Item
          name: CylinderTunnel
          colors:
          - !RGB {r: 204, g: 0, b: 204 }
          - !RGB {r: 204, g: 0, b: 204 }
          - !RGB {r: 204, g: 0, b: 204 }
        - !Item
          name: GoodGoal

\end{minted}
\vspace{-10pt}
\caption{Yaml file for figure \ref{fig:example_config}}
\label{alg:example}
\end{yaml}
\end{minipage}

First of all, this arena has a time step limit defined by \texttt{t: 600} timesteps. If this was set to $0$ the episode would only end once a reward is collected. We then define a list of items, three in this case: \textit{Cube}, \textit{CylinderTunnel} and \textit{GoodGoal}. For each item we can specify lists of values for \texttt{positions}, \texttt{rotations}, \texttt{colors} and \texttt{sizes}. If none of these is specified all the characteristics of the item will be randomised if they can be. For example the \textit{GoodGoal} above will have a random location, size and rotation, but will be green as this is always true. 

For a given item, the number of instances we will attempt to spawn is equal to $max(|$\texttt{positions}$|, |$\texttt{colors}$|, |$\texttt{sizes|}$, |$\texttt{rotations}$|)$ where $|\cdot|$ is the length of a list. For example above we will attempt to spawn two \textit{Wall} and three \textit{CylinderTunnel}. We use a simple \texttt{Vector3} to provide values for \texttt{positions}, \texttt{sizes} which can contain floats, and we use a similar \texttt{RGB} structure for \texttt{colors} which can contain int represent RGB values. Any of these values can be randomised by providing a value of $-1$, as for the second wall which has a randomised \texttt{x} coordinate. Floats in $[0,360]$ can be used for the list \texttt{rotations}, again, a value of $-1$ will randomise the rotation. 

Finally the \texttt{blackouts: [5,10,15,20,25]} let the user define steps at which the light should be turned on or off, providing black pixels to the agent when the light is off. In this case for example the light will be on for the first 5 steps, then off for the next 5, then on for 5 and so on and so forth. Another way to define this is by using a negative value, for example: \texttt{blackouts: [-20]}, which would then turn the light on and off every 20 steps.

\subsection{Items Spawn}

When provided with an arena configuration, the Animal-AI environment will attempt to spawn as many items as possible. However, it does not allow for items colliding at inception. This means that if \texttt{item1} and \texttt{item2} are defined one after the other in the configuration file and that they overlap, only \texttt{item1} will be spawned. One can however spawn objects on top of each other as long as they do not collide at $t=0$, they can of course fall onto each other. 

When spawning items, the environment will start with the items listed first in the configuration file, and move down the list. If any parameter is randomised, the environment will create the object with a certain value for the randomised parameter. It will then attempt to spawn the item where it should appear, if the spot is free the object will spawn there, otherwise the item is deleted, a new one is created with a new value for the randomised parameter and the environment tries again. This loop will go on for 20 iterations in total, if at the end of it no spot has been found for the object, it is ignored and the environment will move on to the next item in the list. This means that for long configuration files that include randomness, the last items in the list are less likely to spawn than the first.

\subsection{Example: Maze Curriculum}

An example application for using such configuration files during training is the PPO agent we trained earlier. We can define a series of YAML files, each of increasing difficulty to solve, and train the agent on each one after the other.

Configurations \ref{alg:curriculum1}, \ref{alg:curriculum2} and \ref{alg:curriculum3} are the configuration files used for arenas shown in figures \ref{fig:curriculum:1}, \ref{fig:curriculum:2} and \ref{fig:curriculum:4} respectively. In our example we created arenas with walls randomly placed along the $x$ and $z$ axes at given values, adding one or two walls per level. We switch from one level of difficulty to the next once the agent reach a certain success rate.

\begin{minipage}[]{\linewidth}
\begin{yaml}
\begin{minted}[
    gobble=4,
    frame=single
  ]{yaml}
    !ArenaConfig
    arenas:
      0: !Arena
        t: 250
        items:
        - !Item
          name: Wall
          positions:
          - !Vector3 {x: -1, y: 0, z: 10}
          rotations: [90]
          sizes:
          - !Vector3 {x: 1, y: 5, z: 9}
        - !Item
          name: GoodGoal
          positions:
            - !Vector3 {x: -1, y: 0, z: 35}
          sizes:
            - !Vector3 {x: 2, y: 2, z: 2}
        - !Item
          name: Agent
          positions:
            - !Vector3 {x: -1, y: 1, z: 5}

\end{minted}
\vspace{-10pt}
\caption{YAML file for figure \ref{fig:curriculum:1}}
\label{alg:curriculum1}
\end{yaml}
\end{minipage}

\vspace{20pt}

\begin{minipage}[]{\linewidth}
\begin{yaml}
\begin{minted}[
    gobble=4,
    frame=single
  ]{yaml}
    !ArenaConfig
    arenas:
      0: !Arena
        t: 400
        items:
        - !Item
          name: Wall
          positions:
          - !Vector3 {x: -1, y: 0, z: 10}
          - !Vector3 {x: -1, y: 0, z: 20}
          - !Vector3 {x: -1, y: 0, z: 30}
          rotations: [90,90,90]
          sizes:
          - !Vector3 {x: 1, y: 5, z: 9}
          - !Vector3 {x: 1, y: 5, z: 9}
          - !Vector3 {x: 1, y: 5, z: 9}
        - !Item
          name: GoodGoal
          positions:
            - !Vector3 {x: -1, y: 0, z: 35}
          sizes:
            - !Vector3 {x: 2, y: 2, z: 2}
        - !Item
          name: Agent
          positions:
            - !Vector3 {x: -1, y: 1, z: 5}

\end{minted}
\vspace{-10pt}
\caption{YAML file for figure \ref{fig:curriculum:2}}
\label{alg:curriculum2}
\end{yaml}
\end{minipage}

\newpage

\begin{minipage}[]{\linewidth}
\begin{yaml}
\begin{minted}[
    gobble=4,
    frame=single
  ]{yaml}

    !ArenaConfig
    arenas:
      0: !Arena
        t: 500
        items:
        - !Item
          name: GoodGoal
          sizes:
            - !Vector3 {x: 2, y: 2, z: 2}
        - !Item
          name: Wall
          positions:
          - !Vector3 {x: -1, y: 0, z: 5}
          - !Vector3 {x: -1, y: 0, z: 10}
          - !Vector3 {x: -1, y: 0, z: 15}
          - !Vector3 {x: -1, y: 0, z: 20}
          - !Vector3 {x: -1, y: 0, z: 25}
          - !Vector3 {x: -1, y: 0, z: 30}
          - !Vector3 {x: -1, y: 0, z: 35}
          - !Vector3 {x: 5, y: 0, z: -1}
          - !Vector3 {x: 10, y: 0, z: -1}
          - !Vector3 {x: 15, y: 0, z: -1}
          - !Vector3 {x: 20, y: 0, z: -1}
          - !Vector3 {x: 25, y: 0, z: -1}
          - !Vector3 {x: 30, y: 0, z: -1}
          - !Vector3 {x: 35, y: 0, z: -1}
          rotations: [90,90,90,90,90,90,
                    90,0,0,0,0,0,0,0]
          sizes:
          - !Vector3 {x: 1, y: 5, z: 9}
          - ... # idem x6
          - !Vector3 {x: 1, y: 5, z: 9}

\end{minted}
\vspace{-10pt}
\caption{YAML file for figure \ref{fig:curriculum:4}}
\label{alg:curriculum3}
\end{yaml}
\end{minipage}

\subsection{Running the Environment}

The Animal-AI environment can be used in two different ways: play mode or training/inference mode. Both modes accept configurations as described in the previous parts as inputs. Play mode allows participants to visualise an arena after defining configuration files in order to assess their correctness. It also permits to control the agent in order to explore and solve the problem at hand.

\subsubsection{Play Mode}

\begin{figure}[h!] 
    \centering
    \begin{subfigure}[b]{0.40\textwidth}
        \includegraphics[width=\textwidth]{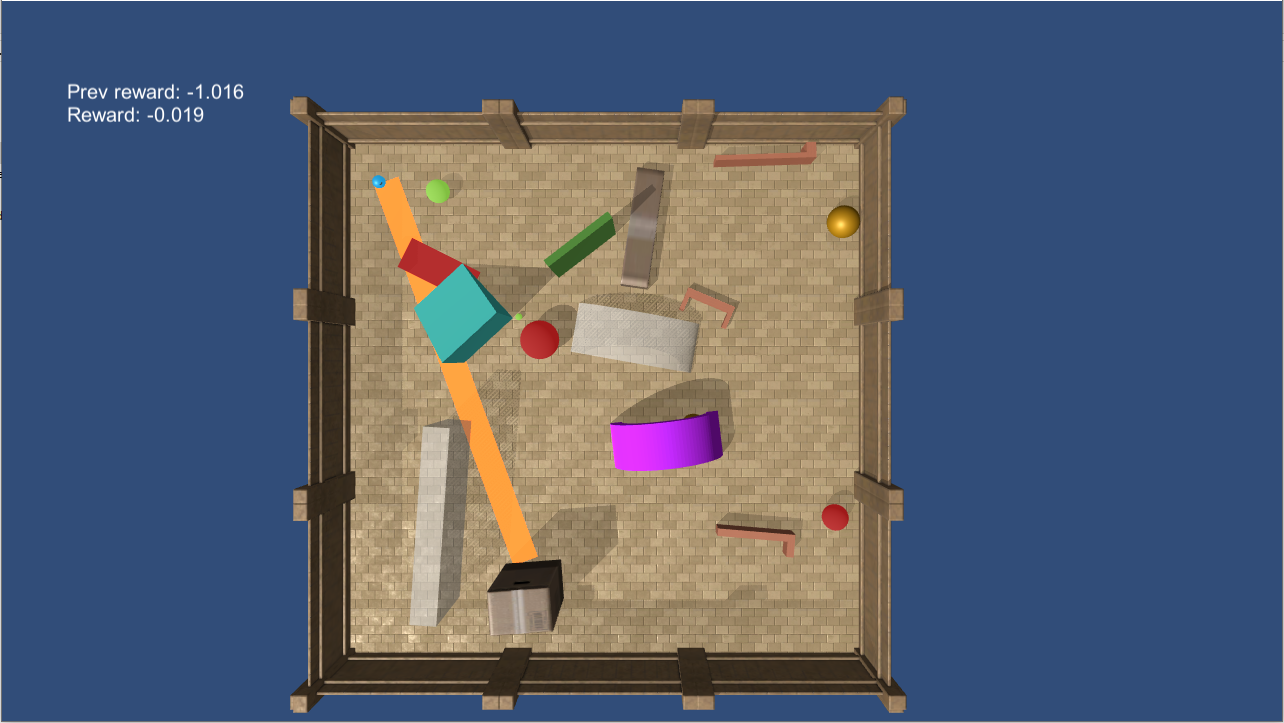}
        \caption{Play mode bird's eye view}
        \label{fig:play:above}
    \end{subfigure}
    \\
    \begin{subfigure}[b]{0.40\textwidth}
        \includegraphics[width=\textwidth]{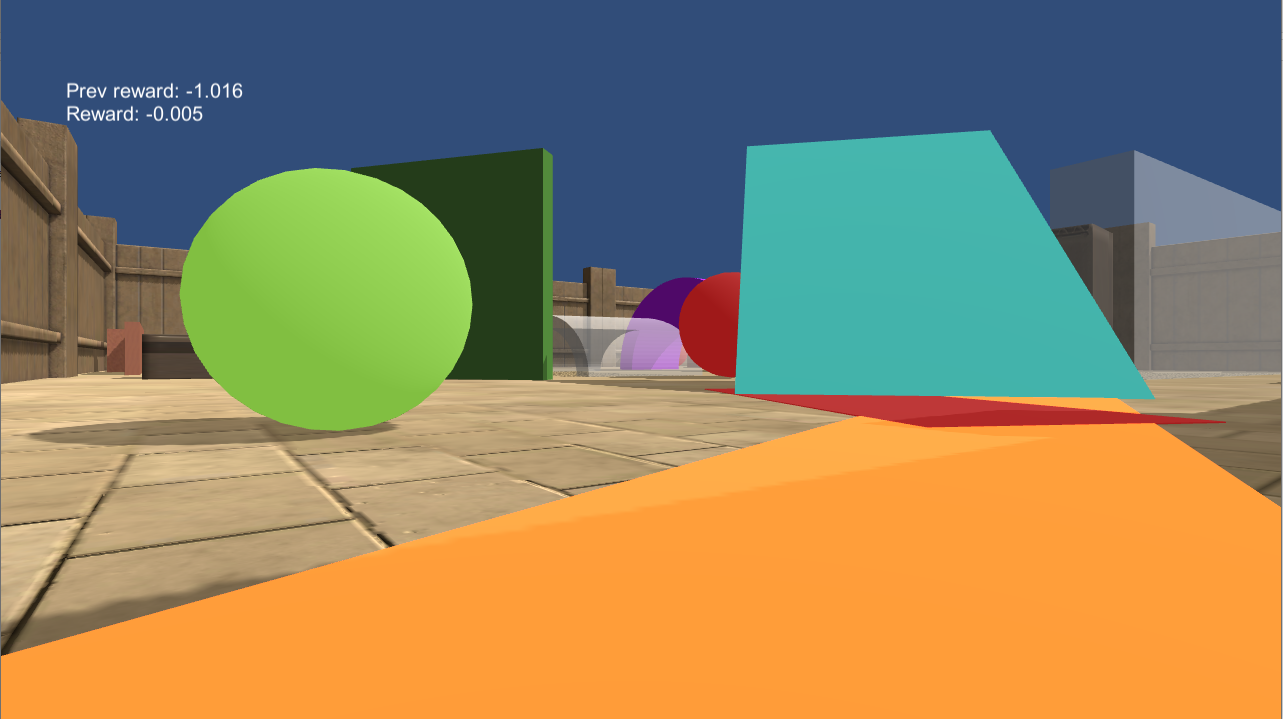}
        \caption{Play mode first person view}
        \label{fig:play:fpv}
    \end{subfigure}
    \caption{Two views from the play mode}
    \label{fig:play}
\end{figure}

Running the executable will start an arena with all the objects cited above in it. As shown in figure \ref{fig:play}, the user can change between a bird's eye view (fig. \ref{fig:play:above}) or a first person view (fig. \ref{fig:play:fpv}). In first person the agent can be controlled using W, A, S, D on the keyboard. Pressing C changes the view and R resets the arena. The score of the previous and current episodes are shown on the top left corner.

\subsubsection{Training and Inference Mode}

To interface with the environment we provide two similar ways: (i) a classic OpenAI Gym API and (2) a Unity ML-Agents API. Both contain the usual functions: \texttt{reset} which takes as optional argument a configuration as defined earlier, and \texttt{step} which sends actions to the environment and returns observations, rewards and additional information needed for training and inference. More details are provided on the online documentation.

\subsection{Trained PPO parameters}

For the experiment presented in the main paper, we used the PPO implementation provided with ML Agents, using the following hyper-parameters:

\begin{itemize}
    \item epsilon: $0.2$
    \item gamma: $0.99$
    \item lambda: $0.95$
    \item learning rate: $3.0e^{-4}$
    \item memory size: $256$
    \item normalise: false
    \item sequence length: $64$
    \item summary freq: $1000$
    \item use recurrent: false
    \item use curiosity: true
    \item curiosity strength: $0.01$
    \item curiosity enc size: $256$
    \item time horizon: $128$
    \item batch size: $64$
    \item buffer size: $2024$
    \item hidden units: $256$
    \item num layers: $1$
    \item beta: $1.0e^{-2}$
    \item max steps: $10e^6$
    \item num epoch: $3$
\end{itemize}

\end{document}